\documentclass[letterpaper, 10 pt, conference]{ieeeconf}  % Comment this line out if you need a4paper

\IEEEoverridecommandlockouts                              % This command is only needed if 
\usepackage{graphicx}
\usepackage{amsmath} % assumes amsmath package installed
\usepackage{amssymb} % assumes amsmath package installed
\usepackage{subfig}
\usepackage{float}
\let\subparagraph\paragraph
\usepackage{titlesec}
\usepackage{bbm}
\usepackage{cuted}
\usepackage[ruled,vlined,linesnumbered]{algorithm2e}
\usepackage{booktabs}
\usepackage{multirow}

\makeatletter
\let\NAT@parse\undefined
\makeatother

\usepackage[breaklinks=true,colorlinks,bookmarks=false]{hyperref}
\usepackage[noabbrev,capitalise]{cleveref}

\usepackage{url}
\usepackage{color}

\newcounter{example}[section]

\titlespacing*{\section}{0pt}{1.5mm}{1.5mm}
\titlespacing*{\subsection}{0pt}{.75mm}{.75mm}
\titlespacing*{\subsubsection}{0pt}{1mm}{1mm}
\newcommand{\xt}[1]{z_{#1}}
\newcommand{\ut}[1]{u_{#1}}
\newcommand{\dt}[1]{d_{#1}}

\newcommand{\dataset}[1]{\mathcal{S}_{{#1}}}

\newcommand{\gtn}[2]{g_\mathrm{#1}^{#2}}

\renewcommand{\baselinestretch}{0.92}

\title{\LARGE \bf
Semi-Supervised Trajectory-Feedback Controller Synthesis\\for Signal Temporal Logic Specifications
}

\author{Karen Leung, Marco Pavone% <-this % stops a space
\thanks{The NASA University Leadership Initiative (grant \#80NSSC20M0163) and Toyota Research Institute (TRI) provided funds to support this work.}%
\thanks{Department of Aeronautics and Astronautics, Stanford University.
        {\tt\small  \{karenl7, pavone\}@stanford.edu}}}%

\begin{document}

\maketitle
\thispagestyle{empty}
\pagestyle{empty}

%%%%%%%%%%%%%%%%%%%%%%%%%%%%%%%%%%%%%%%%%%%%%%%%%%%%%%%%%%%%%%%%%%%%%%%%%%%%%%%%
\begin{abstract}
There are spatio-temporal rules that dictate how robots should operate in complex environments, e.g., road rules govern how (self-driving) vehicles should behave on the road.
However, seamlessly incorporating such rules into a robot control policy remains challenging especially for real-time applications.
In this work, given a desired spatio-temporal specification expressed in the Signal Temporal Logic (STL) language, we propose a semi-supervised controller synthesis technique that is attuned to human-like behaviors while satisfying desired STL specifications. 
Offline, we synthesize a trajectory-feedback neural network controller via an adversarial training scheme that summarizes past spatio-temporal behaviors when computing controls, and then online, we perform gradient steps to improve specification satisfaction.
Central to the offline phase is an imitation-based regularization component that fosters better policy exploration and helps induce naturalistic human behaviors. Our experiments demonstrate that having imitation-based regularization leads to higher qualitative and quantitative performance compared to optimizing an STL objective only as done in prior work.
We demonstrate the efficacy of our approach with an illustrative case study and show that our proposed controller outperforms a state-of-the-art shooting method in both performance and computation time.
\end{abstract}

\section{Introduction}\label{sec:intro}
As robots begin to operate in novel and complex environments (e.g., autonomous driving, human-robot manipulation tasks), there is often underlying structure, or rules, that restrict how a robot should behave. For instance, there are road rules that govern the motion of (self-driving) vehicles on the road.
A core challenge is in seamlessly incorporating the structural information into a robot autonomy stack while respecting other performance metrics, such as intuitive and naturalistic behaviors.
In this work, we develop a learning-based controller synthesis technique that leverages heterogeneous structure, namely structure stemming from temporal logic and expert demonstrations to (i) instill robot behaviors that satisfy desired high-level specifications while attuned to human intuition, and (ii) improve exploration of the search space during the control synthesis process.
In essence, we show that solely optimizing for rule satisfaction can lead to sub-optimal behaviors, and instead, incorporating a few human demonstrations can easily provide significant improvements especially in data scarce regimes.

A common way to incorporate desired high-level specifications into a planner and/or controller is to use temporal logic languages, i.e., formal languages designed to reason about propositions qualified in terms of time. 
In particular, Linear Temporal Logic (LTL) \cite{BaierKatoen2008} and more recently Signal Temporal Logic (STL) \cite{MalerNickovic2004} are popular languages used in robotics. These languages can translate specifications into a mathematical representation for which there are techniques to synthesize planning and control algorithms, e.g., \cite{FainekosGirardEtAl2008,WolffTopcuEtAl2013,KaramanSanfeliceEtAl2008,RamanDonzeEtAl2014}.
In contrast to LTL which is defined over atomic propositions (i.e., discrete states), STL is defined over continuous real-valued signals and encompasses a notion of \emph{robustness}, a scalar measuring the degree of specification satisfaction/violation. Accordingly, there has been a growing interest in using STL robustness in gradient-based methods for controller synthesis (e.g., \cite{PantAbbasEtAl2017,InnesRamamoorthy2020,YaghoubiFainekos2019,LiuMehdipourEtAl2021}).
Recently, \texttt{stlcg} \cite{LeungArechigaEtAl2020}, a toolbox leveraging Pytorch \cite{PaszkeGrossEtAl2017} to compute STL robustness, was developed. As such, \texttt{stlcg} bridges the gap between temporal logic and deep learning through a common computational backbone and therefore provides a natural way to combine temporal logic with deep learning.

\begin{figure}[t]
    \centering
    \includegraphics[width=0.45\textwidth]{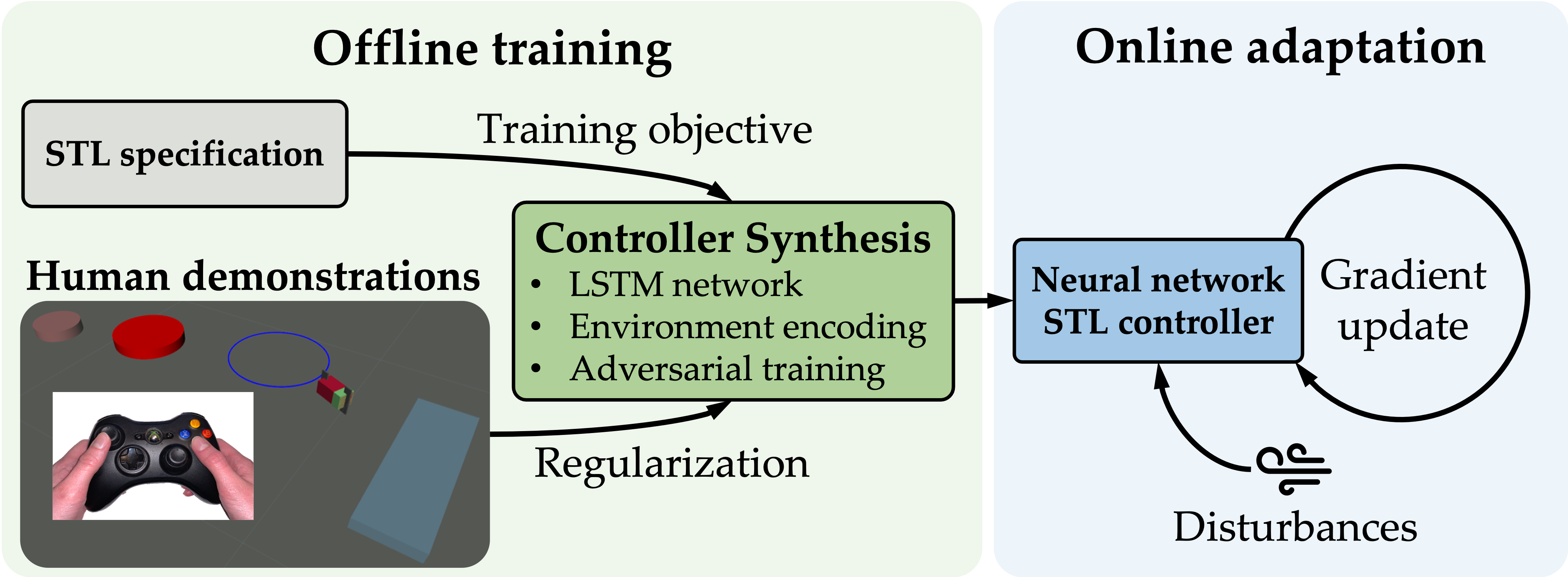}
    \caption{We use Signal Temporal Logic (STL) and expert demonstrations to synthesize a trajectory-feedback controller that satisfies a desired spatio-temporal specification while staying consistent with intuitive human-like behaviors. Offline, an LSTM network and an environment encoding are optimized via an adversarial training strategy.
    Online, a gradient-based adaptation scheme refines the controller to further robustify against disturbances.}
    \label{fig:hero}
\end{figure}

In this paper, we leverage heterogeneous structure, namely STL and expert demonstrations, to synthesize a neural network robot controller, and show that using heterogeneous structure fosters better exploration when searching over the space of controller parameters and therefore leads to higher performing and more intuitive robot behaviors.
At the same time, we are cognizant of the difficulties neural network verification \cite{LiuArnonEtAl2021}.
Thus key to this endeavor, we propose a complementary online adaptation scheme that updates the controller to help robustify against disturbances and protect against limitations in the pre-computed controller.
Figure~\ref{fig:hero} illustrates this two-phased controller synthesis pipeline.

\noindent \emph{Contributions}: Our contributions are fourfold: 
{\bf (i)} We develop a semi-supervised controller synthesis method designed to satisfy a desired STL specification and demonstrate the benefit of using (few) expert demonstrations to help guide the synthesis process.
Crucially, we synthesize a \emph{trajectory-feedback} controller since satisfaction of an STL specification is history-dependent.
{\bf (ii)} We generalize our trajectory-feedback controller to new but similarly structured environments to prevent re-synthesizing a new STL controller whenever the environment changes (e.g., obstacles move). Environment generalization is achieved by conditioning the control parameters on environment parameters.
{\bf (iii)} We combine an offline iterative adversarial training algorithm with an online adaptation scheme to improve robustness against disturbances. We show empirically that even when the neural network controller (trained offline) produces trajectories that violate the STL specification, the online adaptation step results in satisfying trajectories.
{\bf (iv)} We demonstrate our controller on a relatively complex STL specification and show that it outperforms a state-of-the-art shooting method in terms of STL robustness and computation time.

\section{Related Work}\label{sec:related_work}
We provide an overview of state-of-the-art temporal logic control synthesis methods, and learning-based controllers that use temporal logic.

\subsection{Temporal Logic Control Synthesis}
\label{subsec:related_work_temporal_logic_synthesis}
Temporal logic provides a formalism to express specifications in natural language into a concise mathematical representation.
A popular temporal language is LTL which defined over atomic propositions (i.e., discrete states) and there are well-studied automaton-based methods for synthesizing correct-by-construction \emph{closed-loop} controllers \cite{ClarkeGrumbergEtAl1999} satisfying LTL specifications.
However, the synthesis procedure is doubly exponential \cite{PneuliRosner1989} and therefore a smaller fragment of the LTL language is used instead.
On the other hand, STL is more expressive than LTL and is defined over continuous real-valued signals.
Unfortunately, analogous controller synthesis approaches for STL do not exist, and the design of such methods remain an open problem.

Instead of synthesizing a closed-loop controller, an open-loop trajectory satisfying an LTL or STL specification can be constructed by solving a Mixed Integer Linear Program (MILP) and executed in a receding horizon fashion \cite{KaramanSanfeliceEtAl2008,RamanDonzeEtAl2014,WolffTopcuEtAl2014,SadraddiniBelta2015,RamanDonzeEtAl2015,SusmitRajEtAl2018}.
% The MILP formulation can account for other control objectives such as control effort, and is applicable to any LTL specifications. 
While receding horizon control can adapt to environment changes, MILPs are NP-hard and do not scale well with specification complexity and trajectory length.
As such, MILP-based approaches may become impractical for general nonlinear systems that need to perform complex tasks over long time horizons.
Instead, \cite{PantAbbasEtAl2017} utilizes smooth approximations of STL robustness formulas to design a sequential quadratic program (SQP) to compute controls that maximize robustness. While an SQP can account for nonlinear dynamics, the solve time is still intractable for real-time applications. Additionally, the solution may converge to an undesirable local minimum.
To address the high computation times, \cite{PantAbbasEtAl2018} proposes a hierarchical approach tailored for reach-avoid multi-quadrotor missions. The algorithm optimizes over sparse way-points used to generate a higher-resolution continuous-time minimum jerk trajectory. While hardware experiments seem promising, the computation solve time was still a bottleneck especially when scaling up to more agents and increasing the time horizon. 

In summary, STL controller synthesis is still a challenging problem, and computational tractability is a large hurdle especially for problems with long horizons and complex STL specifications.

\subsection{Temporal Logic in Learning-based Approaches}
Recently, there has been a growing interest in using STL as a form of inductive bias within a variety of learning-based approaches, such as in deep neural networks \cite{LeungArechigaEtAl2020,LiRosmanEtAl2020}, reinforcement learning \cite{LiVasileEtAl2017,JiangBharadwajEtAl2021}, and learning from demonstrations \cite{InnesRamamoorthy2020}. 
Given an STL specification, these approaches augment the loss (or reward) with an STL robustness term.
Deep neural networks provide a computationally tractable way to synthesize STL controllers for complex systems.
However, it is difficult to formally verify that the resulting neural network will satisfy the desired (spatio-temporal) specification for all possible inputs \cite{LiuArnonEtAl2021}. 

A common approach to bolster neural network controller performance is to leverage an adversarial training step where a search procures falsifying samples to be used for retraining the network (e.g., \cite{InnesRamamoorthy2020,YaghoubiFainekos2019,LandryDaiEtAl2021}). The process is repeated until no more falsifying samples can be found or a stopping criteria is met. While this generally improves the performance of a model, it does not necessarily provide any formal guarantees on performance.

Works most similar to this paper are \cite{YaghoubiFainekos2019} and \cite{LiuMehdipourEtAl2021}.
In \cite{YaghoubiFainekos2019}, a state-feedback feedforward neural network controller was synthesized via an iterative adversarial training scheme whereby the training objective maximized STL robustness only. However, the state-feedback element of the controller prevents exploiting knowledge of past spatio-temporal behaviors, therefore restricting the types of specifications that are applicable. 
% In general, satisfaction of an STL specification depends on the \emph{entire} trajectory (i.e., past and future trajectory), and not just a single state.
Additionally, the method is tailored towards a fixed environment, thus requiring re-synthesis if the environment changes.
Instead, \cite{LiuMehdipourEtAl2021} synthesizes a \emph{recurrent} neural network (RNN) controller to account for past spatio-temporal behaviors, and then online, the controller is complemented with a control barrier function \cite{AmesCooganEtAl2019} to avoid collision with new unseen obstacles. However, the approach (i) focuses on simple reach-avoid STL specifications, (ii) environment variation is limited to obstacles only, and (iii) is a fully supervised approach that assumes access to a large data set of STL-satisfying trajectories generated by solving an STL-constrained trajectory optimization problem. As discussed previously, solving an STL-constrained trajectory optimization problem is nontrivial even if performed offline.

In summary, learning-based techniques provide a more computationally tractable approach for synthesizing closed-loop STL controllers, but, unfortunately, they alone lack strict guarantees on specification satisfaction.
In this work, we develop a semi-supervised learning-based controller synthesis method that produces an STL trajectory-feedback controller deployable in a variety of environments. To address the limitations in learning-based controllers, we additionally propose an online adaptation scheme to correct for any STL violation.
% Specifically, we learn offline a closed-loop controller and adapt the controller online to account for disturbances.
Compared to similar works, we demonstrate that our method is applicable to more complex STL specifications, and is more data efficient.

\section{Signal Temporal Logic}\label{sec:stl}

In this section, we review the definitions and syntax of STL, and introduce the quantitative semantics which are used to compute robustness. 
STL formulas are interpreted over signals, $s_{t} = \xt{t}, \xt{t+1},...,\xt{t+T}$, an ordered finite sequence of states $\xt{i}\in\mathbb{R}^n$. 
% For ease of notation, $s$ (i.e., when the subscript on $s_t$ is dropped) denotes the entire signal.
A signal represents a sequence of real-valued, discrete-time outputs from any system of interest. In this work, we assume that a signal is sampled at uniform time steps.
STL formulas are defined recursively according to the following grammar (written in Backus-Naur form),
\begin{align}
\phi::= &~~\top ~|~ \mu_c ~|~ \neg\phi ~|~ \phi \wedge \psi ~|~  \phi\,\mathcal{U}_{[a,b]}\,\psi,
\label{eq:STL grammar}
\end{align}
\noindent where $\top$ means true, $\mu_c$ is a predicate of the form $\mu(z) > c$, where $c\in\mathbb{R}$ and $\mu: \mathbb{R}^n \rightarrow \mathbb{R}$ is a differentiable function, $\phi$ and $\psi$ are STL formulas, and $[a,b]\subseteq \mathbb{R}_{\geq 0}$ is a time interval. 
When the time interval is omitted, the temporal operator is evaluated over the positive ray $[0, \infty )$.
The symbols $\neg$ (negation/not), and $\wedge$ (conjunction/and) are logical connectives, and $\mathcal{U}$ (until) is a temporal operator.
Additionally, other commonly used logical connectives ($\vee$ (disjunction/or) and $\Rightarrow$ (implies)), and temporal operators ($\lozenge$ (eventually), and $\square$ (always)) can be derived from \eqref{eq:STL grammar}.
% A predicate is an STL formula. If $\phi$ and $\psi$ are STL formulas, then applying a logical connective or temporal operator produces another STL formula, i.e, $\varphi = \phi \wedge \psi$ is also an STL formula.

% Given a signal and an STL formula $\phi$, we can use Boolean semantics to check if the signal satisfies $\phi$. 
We use the notation $s_t \models \phi$ to denote that a signal $s_t$ satisfies an STL formula
$\phi$. For brevity, we omit the Boolean semantics (see \cite{LeungArechigaEtAl2020} for details) and instead describe the temporal operators informally. \textbf{Until}: $s_t \models \phi \,\mathcal{U}_{[a,b]} \psi$ if there
is a time $t^\prime \in [t + a, t + b]$ such that $\phi$ holds for all time before $t^\prime$ and $\psi$ holds
at time $t^\prime$.
\textbf{Eventually}: $s_t \models \lozenge_{[a,b]} \phi$ if at some time $t^\prime \in [t + a, t + b]$, $\phi$ holds at least once. \textbf{Always}: $s_t \models \square_{[a,b]} \phi$ if $\phi$ holds for all $t^\prime \in [t + a, t + b]$.

Further, STL admits a notion of robustness. That is, there are quantitative semantics that measure the degree of satisfaction (positive robustness value) or violation (negative robustness value) of an STL formula given a signal. The quantitative semantics are defined as follows,
% For brevity and for the purposes of aiding discussion in future sections, we will present the robustness formulas, though for more details, refer to \cite{LeungArechigaEtAl2020}.

{\small
\begin{equation*}
\begin{split}
\rho_\top(s_t) & \:= \: \rho_{\max} \: \text{  where $\rho_{\max} > 0$}\\
\rho_{\mu_c}(s_t) & \:= \: \mu(\xt{t}) - c, \quad \quad  \rho_{\neg \phi}(s_t) \:= \: -\rho_\phi(s_t)\\
\rho_{\phi \wedge \psi}(s_t) & \:= \: \min(\rho_\phi(s_t), \rho_\psi(s_t))\\
\rho_{\phi \vee \psi}(s_t ) & \:= \: \max(\rho_\phi(s_t), \rho_\psi(s_t))\\
\rho_{\phi \Rightarrow \psi}(s_t) & \:= \: \max(-\rho_\phi(s_t), \rho_\psi(s_t))\\
\rho_{\lozenge_{[a,b]} \phi}(s_t ) & \:= \: \max_{t^\prime \in [t+a, t+b]}\rho_\phi(s_{t^\prime})\\
\rho_{ \square_{[a,b]} \phi}(s_t) & \:= \: \min_{t^\prime \in [t+a, t+b]}\rho_\phi(s_{t^\prime})\\
\rho_{\phi \,\mathcal{U}_{[a,b]}\,\psi}(s_t) & \: = \max_{t^\prime \in [t+a, t+b]}\left[\min\left(\rho_\psi(s_{t^\prime}), \, \min_{t^{\prime\prime}\in [t, t^\prime]} \rho_\phi(s_{t^{\prime\prime}})\right)\right].
\end{split}
\end{equation*}
}
\noindent Using these robustness formulas, we can compute gradients of STL robustness with respect to the input signal \cite{LeungArechigaEtAl2020}.

\begin{figure*}[h]
    \centering
    \includegraphics[width=\textwidth]{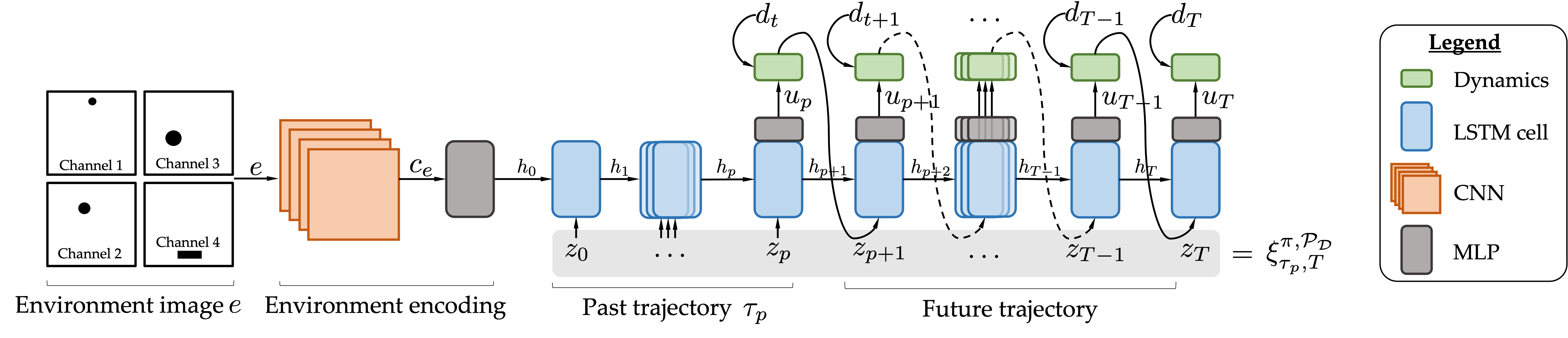}
    \caption{Neural network architecture of a trajectory-feedback STL controller. The environment image is used in computing the LSTM initial hidden state via CNN and MLP layers. The past trajectory is passed through the LSTM cell and then future states and controls are generated in an auto-regressive manner using the LSTM cell and integration through the system dynamics (with disturbance $\dt{t}$). Note: The dynamics depend on state but the arrows are omitted to reduce visual clutter.}
    \label{fig:architecture}
\end{figure*}

\section{Problem Formulation}\label{sec:prob_form}
% find a policy pi so that you satisfy the STL formula
Let $\xt{t} \in \mathcal{X} \subset \mathbb{R}^n$, $\ut{t} \in \mathcal{U} \subset \mathbb{R}^m$, $\dt{t} \in \mathcal{D} \subset \mathbb{R}^{m_d}$ be the state, control, and disturbance of a system at time $t$ respectively. Let $\tau_p = \xt{0:p}$ denote a state trajectory from timestep $0$ to $p$.
Let $\mathcal{X}_0 \subseteq \mathcal{X}$ be the set of states a system starts in.
Further, let the time-invariant, discrete-time state space dynamics for a system be $\xt{t+1} = f(\xt{t}, \ut{t}, \dt{t})$, and $e\in \mathcal{E}$ denote the set of environment parameters (e.g., image of the environment) that a system operates in.
Let $\xi_{\tau_p, T}^{\pi, \mathcal{P}_\mathcal{D}} = \xt{0:T}$ denote a trajectory where $\xt{0:p} = \tau_p$, and $\xt{p+1:T}$ is produced by the dynamics $f(\cdot, \cdot, \cdot)$ following a control policy $\ut{t} = \pi(\cdot)$ subject to a stochastic disturbance $\dt{t}\sim \mathcal{P}_\mathcal{D}$ at each time step. For ease of notation, when $p=0$, we write $\xi_{\xt{0}, T}^{\pi, \mathcal{P}_\mathcal{D}}$.
Let $\phi$ represent an STL specification that we desire a system to satisfy. Then the problem we seek to solve is:

\noindent {\bf STL controller synthesis problem}:
\textit{For a time horizon $T$, find $\pi(\cdot)$ such that with $\dt{t} \sim \mathcal{P}_\mathcal{D}$,  $\forall e\in \mathcal{E}$, and $\forall \xt{0}\in\mathcal{X}_0$,  $\xi_{\xt{0}, T}^{\pi, \mathcal{P}_\mathcal{D}} \models \phi$. 
In words, we want to find a control policy $\pi$ such that under disturbance inputs $\dt{t} \sim \mathcal{P}_\mathcal{D}$, for all possible environments in $\mathcal{E}$, and initial states in $\mathcal{X}_0$, all trajectories $\xi_{\xt{0}, T}^{\pi, \mathcal{P}_\mathcal{D}}$ satisfy $\phi$.}

Unfortunately, solving the STL controller synthesis problem exactly is challenging; the disturbance, nonlinearity, and recursiveness of robustness formulas make finding a globally optimal solution difficult. Additionally, STL satisfaction depends on past and future trajectories.

\section{STL Control Synthesis}
We propose a learning-based controller synthesis framework that (i) leverages expert demonstrations to aid policy exploration and induce intuitive behaviors, (ii) uses an adversarial training scheme to improve the closed-loop policy, and (iii) employs an online adaptation step for added robustification against disturbances.

\subsection{Overview}
Our method represents a \emph{middle ground between receding-horizon open-loop control and closed-loop control}. 
In the offline computation, we construct a data-driven closed-loop trajectory-feedback controller via an iterative adversarial training scheme. However, due to limitations in neural network verification, the resulting controller may result in violating trajectories for some initial states, environment, and disturbance inputs. To address this limitation, a lightweight online computation will update the controller whenever a falsifying trajectory is expected to occur. This is in contrast to receding horizon optimal control methods that solve a potentially costly optimization problem at each time step.

\subsection{Trajectory-feedback Controller Architecture}\label{subsec:architecture}
Consideration of past spatio-temporal behaviors is critical when reasoning about the spatio-temporal properties of an \emph{entire} signal. Therefore we use a Long Short Term Memory (LSTM) \cite{HochreiterSchmidhuber1997,GersSchmidhuberEtAl1999} network, a specific type of RNN, to construct a \emph{trajectory-feedback} controller. 
To generalize to different environments, we condition the neural network controller on environment parameters via the LSTM initial hidden state. 
% Since $\mathcal{E}$ represents the set of parameters describing the environment, we use $e\in\mathcal{E}$ to construct an LSTM initial hidden state.
In this work, we assume that $\mathcal{E}$ is a set of images describing possible layouts of the environment, and therefore we use a Convolutional Neural Network (CNN) \cite{LeCunBengioEtAl2015} to summarize $e\in\mathcal{E}$.
However, different transformations may be used depending on the environment representation (e.g., occupancy grid, vector of parameter values).
Figure \ref{fig:architecture} illustrates a schematic of the proposed neural network architecture.
Since the policy $\pi$ depends on the past trajectory and the environment, we write $\ut{t} = \pi_\theta(\tau_t, e)$ where $\theta$ denotes a vector of neural network parameters.

\subsubsection{LSTM: Trajectory-feedback controller}\label{subsubsec:LSTM}
LSTMs take as input time-series data, and output time-series data.
Using the past trajectory $\tau_t = \xt{0:t}$ that a system has already traversed, an LSTM can summarize, with a hidden state $h_t$, past spatio-temporal behaviors without the need for state augmentation. 
Specifically, let $o_t, h_{t+1} = \gtn{LSTM}{n_h}(\xt{t}, h_t)$ denote the input-output relationship described by an LSTM cell with hidden state size $n_h$ and $o_t, h_t \in \mathbb{R}^{n_h}$.\footnote{For brevity, we omit the details of the internal operations of the LSTM cell. See \cite{HochreiterSchmidhuber1997,GersSchmidhuberEtAl1999} for details on the architecture of the LSTM cell.}$^,$\footnote{For LSTMs, the hidden state $h_t$ is actually a tuple of two vectors, each of length $n_h$.}
At time $t$, a state $\xt{t}$ and hidden state $h_t$ are passed into $\gtn{LSTM}{n_h}$
% : \mathrm{R}^n \times \mathrm{R}^{n_h} \rightarrow \mathrm{R}^{n_h} \times \mathrm{R}^{n_h}$ 
to produce an output state $o_t$, and the next hidden state $h_{t+1}$.
When unrolling the LSTM with the input trajectory $\tau_p$, we simply feed in $\xt{t}$ and $h_t$ sequentially at each time step up until we obtain $h_{p+1}$ and $o_p$. 

For $t \geq p$, the future trajectory is generated in an autoregressive fashion. That is, the output state, $o_t$, is passed through a multi-layer perceptron (MLP), denoted by $\gtn{MLP}{n_h \rightarrow m}: \mathbb{R}^{n_h} \rightarrow \mathbb{R}^m$ which transforms $o_t$ into control inputs.
% : \mathbb{R}^{n_h} \rightarrow \mathbb{R}^m$
% where $m$ is the dimension of $\mathcal{U}$. 
To ensure the control inputs satisfy control constraints $u\in[\underline{u}, \overline{u}]$, we take $\ut{t}^\prime = \gtn{MLP}{n_h \rightarrow m}(o_t)$ and apply the following transformation, $\ut{t} = \frac{\overline{u} - \underline{u}}{2}\tanh{(\ut{t}^\prime)} +  \frac{\overline{u} + \underline{u}}{2}$.
Given the newly computed $\ut{t}$, the current state $\xt{t}$, and disturbance $\dt{t}\sim\mathcal{P}_\mathcal{D}$, the next state can be computed using the dynamics model $\xt{t+1} = f(\xt{t}, \ut{t}, \dt{t})$. 
By incorporating the dynamics in the unrolling of the LSTM, we ensure that the resulting trajectory is dynamically feasible.
The next state $\xt{t+1}$ and next hidden state $h_{t+1}$ are then passed through the LSTM cell again to compute the next output state and control and so forth.
We continue unrolling the LSTM cell up to $t=T$, a predetermined horizon length. The overall trajectory (joining the past trajectory $\tau_p$ and propagated trajectory $\xt{p+1:T}$) is denoted by $\xi_{\tau_p, T}^{\pi_\theta, d}$.
Next, we discuss how to initialize the hidden state of the LSTM network.

\subsubsection{CNN: Environment generalization}\label{subsubsec:CNN}
To generalize the controller to new environments without the need for re-synthesis, we condition the initial LSTM hidden state with an environment summary vector. In this work, we summarize an image of the environment using CNNs, though a different transformation could be used depending on the environment representation.

Given an STL specification $\phi$, there are components within $\phi$ that reference different regions in the environment and describe how the system should interact with those regions.
We propose corresponding each image channel of $e$ to a particular region type. For example, consider $\phi = \lozenge \, \phi_\mathrm{goal} \; \wedge \; \square \neg \phi_\mathrm{obs}$ where $\phi_\mathrm{goal}$ and $\phi_\mathrm{obs}$ are predicates describing being inside the goal and obstacle region respectively. The formula $\phi$ translates to ``eventually reach the goal region and always avoid the obstacle region.'' 
As such, specific to $\phi$, there are elements in the environment that correspond to the goal and obstacle regions (and starting regions too). Thus we make each image channel of $e$ correspond to an image of each region type situated in the environment. For example, the first channel describes an image (matrix of 1 and 0's) of the goal region. The second channel describes an image of the obstacle region, and so forth. See Figure~\ref{fig:architecture} for an example visualization.
We then use a CNN to encode $e$ into a summary vector $c_e$ which is then used to initialize the hidden state of the LSTM.
Let $\gtn{CNN}{|e| \rightarrow n_c}$ be a CNN network encoding the environment image $e$ into a hidden state $c_e \in \mathbb{R}^{n_c}$, and $\gtn{MLP}{n_c \rightarrow n_h}: \mathbb{R}^{n_c} \rightarrow \mathbb{R}^{n_h}$ be an MLP transforming $c_e$ to $h_0$.\footnote{Two MLPs are actually needed since the hidden state of LSTMs is a tuple of two vectors, each of size $n_h$.} Then, the initial hidden state for the LSTM can be computed by, $h_0 = \gtn{MLP}{n_c \rightarrow n_h}(\gtn{CNN}{|e| \rightarrow n_c}(e))$.
Although the structure of $e$ is dependent on the STL specification of interest, we can still generalize across new and unseen environments for which $\phi$ is still valid. For example, a valid environment is one where the obstacle and goal regions change location (we simply use the corresponding $e$ to compute a new $h_0$), but an invalid one would be if the goal region disappeared. 

\subsection{Learning the Control Parameters}\label{subsec:training}

Given the neural network architecture described in Section \ref{subsec:architecture}, the goal is to learn $\theta$, a vector of neural network parameters from the CNN, LSTM, and MLP networks, such that given any $e\in\mathcal{E}$ and any $\xt{0}\in\mathcal{X}_0,\, \xi_{\xt{0}, T}^{\pi_\theta, \mathcal{P}_\mathcal{D}}$, satisfies $\phi$.
There are two key aspects to our training scheme: (i) We leverage expert demonstrations for an imitation regularization loss to help guide the training to a better optimum, compared to the case when optimizing only for STL robustness. For this reason, we refer to our approach as semi-supervised. (ii) We take on an adversarial training approach that iterates between a training step to optimize $\theta$ via gradient descent, and an adversarial step which searches for initial states and environments where the controller produces a violating trajectory. The next training step updates the model using the violating samples.
The pseudocode for the training process is outlined in Algorithm \ref{alg:planner}, and the details are provided next.

We note, however, there are no theoretical guarantees that Algorithm \ref{alg:planner} will converge and that no adversarial samples exist.
We address this limitation in Section \ref{subsec:deploy} by proposing an online adaptation step.

\begin{algorithm}[t]
\SetAlgoLined
\KwResult{$\pi_\theta$}
{\bf Initialization}: $\dataset{} = \lbrace (\xt{0,i} , e_i)\rbrace_{i=1,...,N}$ where $(\xt{0,i}, e_i) \sim \mathrm{Uniform}[\mathcal{X}_0 \times \mathcal{E}]$\;
$k \leftarrow 0$\;
{\bf Model training}: Optimize $\theta$ using $N_\mathrm{full}$ training epochs on \eqref{eq:loss_training} over $\dataset{}$\;
 \While{$k < K$}{
    {\bf Adversarial search}: Find $\dataset{\mathrm{adv}} = \lbrace (\xt{0,j}, e_j) \mid \xi_{x_{0,j}, T}^{\pi_\theta, \mathcal{P}_\mathcal{D}} \not\models \phi \rbrace_{i=1,...,N_\mathrm{adv}}$ with $\theta$ fixed (e.g., acceptance-rejection sampling)\;
    {\bf Update initial states}: Re-sample from $\mathrm{Uniform}[\mathcal{X}_0 \times \mathcal{E}]$ to get $\dataset{0} = \lbrace (\xt{0,i}, e_i) \rbrace_{i=1,...,N}$\;
    $\dataset{} \leftarrow \dataset{0} \cup \dataset{\mathrm{adv}}$\;
    {\bf Model training}: Optimize $\theta$ using $N_\mathrm{mini}$ training epochs on \eqref{eq:loss_training} over $\dataset{}$\;
    $k \leftarrow k + 1$\;
 }
 \caption{STL controller synthesis (offline)}\label{alg:planner}
\end{algorithm}

\subsubsection{Training step (Lines 3 and 8 in Algorithm~\ref{alg:planner})}\label{subsubsec:training_step}
Let $\mathcal{S} = \lbrace (\xt{0,i} , e_i)\rbrace_{i=1,...,N}$ represent samples from $\mathcal{X}_0 \times \mathcal{E}$ (the samples can be sampled uniformly), and let $\Xi^\mathrm{exp} = \lbrace \xi_{\xt{0, i}^\mathrm{exp},T_i}^\mathrm{exp}, e_i^\mathrm{exp} \rbrace_{i=1,...,N_\mathrm{exp}}$ represent trajectories corresponding to expert demonstrations that satisfy $\phi$, the STL specification that we aim to design a controller for.
We make an assumption that we have access to expert demonstrations, such as from real-world operations or from simulation. Approximate solutions from direct numerical optimization could be used but may require additional human supervision for refinement.
We note that with any data-driven approaches, obtaining data may be challenging especially for more complex specifications and systems. However, our approach is \emph{semi}-supervised since the demonstrations are used for regularization instead of the main training objective. Therefore we do not require a significant amount of demonstrations compared to fully supervised approaches (e.g., \cite{LiuMehdipourEtAl2021}), and favorably so if data collection is expensive or demonstrations are scarce.

We apply stochastic gradient descent on $\theta$ to minimize the following loss objective,
\begin{eqnarray}
    \mathcal{L}_\mathrm{train}(\theta;\mathcal{S}, \Xi^\mathrm{exp}) =  \mathcal{L}_\mathrm{STL}(\theta; \mathcal{S}) + \gamma \mathcal{L}_\mathrm{imit}(\theta; \Xi^\mathrm{exp}), \label{eq:loss_training}\\
    \mathcal{L}_\mathrm{STL}(\theta; \mathcal{S}) = \frac{1}{|\dataset{}|}\sum_{(\xt{0}, e) \in \mathcal{S}}\mathrm{LeakyReLU}(-\rho_\phi(\xi_{\xt{0}, T}^{\pi_\theta, \mathcal{P}_\mathcal{D}})), \label{eq:loss_STL}\\
    \mathcal{L}_\mathrm{imit}(\theta; \Xi^\mathrm{exp}) = \frac{1}{N_\mathrm{exp}} \sum_{i = 1}^{N_\mathrm{exp} }\frac{1}{T_i} \Delta( \xi_{\xt{0, i}^\mathrm{exp},T_i}^\mathrm{exp},\xi_{\xt{0, i}^\mathrm{exp},T_i}^{\pi_\theta, \mathcal{P}_\mathcal{D}}; e_i^\mathrm{exp}) \label{eq:loss_reconstruction}
\end{eqnarray}
where $\mathrm{LeakyReLU}(x)=\max(0.01x,x)$ and $\Delta(\xi_a, \xi_b; e) = \mathrm{MSE}(\xt{a}, \xt{b}) + \gamma_\mathrm{imit} \mathrm{MSE}(\ut{a}, \ut{b})$ is a weighted sum of the mean-square-error in state and controls between trajectories $\xi_a$ and $\xi_b$ under environment $e$.
Note that \eqref{eq:loss_STL} differs from maximizing robustness as done in \cite{PantAbbasEtAl2017,InnesRamamoorthy2020,YaghoubiFainekos2019}. The LeakyReLU function focuses primarily on minimizing the amount of violation and focuses less on increasing the amount of satisfaction (by a factor of 0.01).
The purpose of \eqref{eq:loss_reconstruction} is to help regularize the training process since \eqref{eq:loss_STL} is nonlinear and non-convex. Simultaneously, \eqref{eq:loss_reconstruction} helps guide the exploration towards regions where the controller produces trajectories consistent with how humans would behave and therefore avoid superfluous or unreasonable trajectories (e.g., taking unnecessary detours but still satisfy $\phi$).

\subsubsection{Adversarial search (Line 5 in Algorithm~\ref{alg:planner})}\label{subsubsec:adversarial}
The training step is optimized using \emph{samples} of initial states and environments.
As such, there could still exist initial states and environments that lead to negative robustness. The goal of the adversarial search is to find a set of initial states and environments $\mathcal{S}^\mathrm{adv} = \lbrace (\xt{0,i}^\mathrm{adv}, e_i^\mathrm{adv})\rbrace_{i=1,...,N_\mathrm{adv}}$ such that the resulting trajectories violate $\phi$.
A number of methods can be used to search for adversarial samples, such as batched (projected) gradient descent on $\xt{0}$ to minimize robustness, cross-entropy method, or simulated annealing \cite{AnnapureddyLiuEtAl2011}. 
Since we strive to find \emph{any} samples that produce a violating trajectory, we opt for a simpler approach of acceptance-rejection sampling whereby we uniformly sample from $\mathcal{X}_0 \times \mathcal{E}$ and reject any samples that produce satisfying trajectories.
We continue sampling until $N_\mathrm{adv}$ adversarial samples are found or a termination criterion is met.
Using the adversarial samples, and newly sampled initial states and environments, we can continue training the model (lines 6--8 in Algorithm~\ref{alg:planner}).

\begin{algorithm}[t]
\SetAlgoLined
\KwResult{$\tau_T$}
\textbf{Initialization: }STL specification $\phi$, initial state $\xt{0}$, environment $e$, time horizon $T$, neural network weights $\theta_0$, step size $\eta$, and maximum number of gradient steps $N_\mathrm{gd}$\;
$\theta \leftarrow [\tilde{\theta}, \hat{\theta}]$ where $\tilde{\theta}$ are neural network parameters of $\gtn{MLP}{n_h \rightarrow m}$, and $\hat{\theta}$ are the remaining neural network parameters (held fixed)\;
$\tau_t \leftarrow \xt{0}$\;

 \For{$t = 0:T$}{

 \If{$\mathbb{E}[ \rho(\xi_{\tau_t, T}^{\pi_\theta, \mathcal{P}_\mathcal{D}}, \phi) < 0$}{
    $j \leftarrow 0$\;
    \While{$(\mathbb{E}[ \rho(\xi_{\tau_t, T}^{\pi_\theta, \mathcal{P}_\mathcal{D}}, \phi) < 0) \wedge (j < N_\mathrm{gd})$}{
            $\tilde{\theta} \leftarrow \tilde{\theta} + \eta\nabla_{\tilde{\theta}} \mathbb{E}_{d\sim \mathcal{P}_\mathcal{D}}[ \rho(\xi_{\tau_t, T}^{\pi_\theta, \mathcal{P}_\mathcal{D}}, \phi)$\;
            $\theta \leftarrow [\tilde{\theta}, \hat{\theta}]$\;
            $j \leftarrow j + 1$\;
        }
    }
    $\ut{t} = \pi_\theta(\tau_t; e)$\;
    $\xt{t+1} = f (\xt{t}, \ut{t}, \dt{t})$ where $\dt{t} \sim \mathcal{P}_\mathcal{D}$\;
    $\tau_{t} \leftarrow \xt{0:t+1}$\;
 }
 \caption{Deploying $\pi_\theta$ (online)}\label{alg:online adaptation}
\end{algorithm}
\subsection{Deploying the Controller Online}
\label{subsec:deploy}

Unfortunately, after running Algorithm~\ref{alg:planner}, there are no guarantees that the resulting controller will produce satisfying trajectories for \emph{all} initial states and environments, especially under the presence of disturbances to the system. 
There is a lot of effort towards neural network verification, though verifying RNNs especially with temporal logic considerations remains challenging.

To address the limitations of the LSTM controller, Algorithm \ref{alg:online adaptation} proposes updating some of the controller parameters online whenever the controller is expected to produce a violating trajectory. That is, we perform gradient steps on $\tilde{\theta}$, the parameters of $\gtn{MLP}{n_h \rightarrow m}$, when needed.
At each time step, we perform a Monte Carlo estimate of the expected robustness value (line 5). If the expected robustness is negative, then we perform at most $N_\mathrm{gd}$ gradient descent steps on $\tilde{\theta}$ to increase the robustness value (lines 7--10).
If desired, a different risk metric (e.g., Value at Risk) could be used instead.
Then the resulting control is passed into the system and a step is taken forward in time (lines 13--15).

\begin{figure*}[t]
    \centering
    \includegraphics[width=\textwidth]{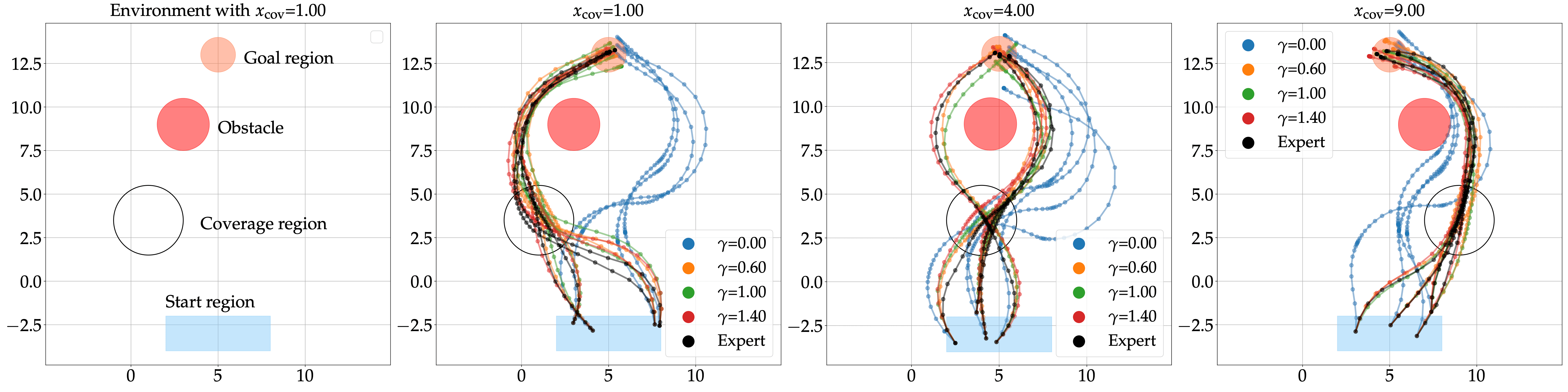}
    \caption{Trajectories generated from models trained with different values of $\gamma$, the imitation regularization weight.}
    \label{fig:human trajectory}
\end{figure*}

\section{Experiments}\label{sec:experiments}
We investigate the offline and online performance of our proposed controller applied to a nonlinear system.

\subsection{Case-Study Set-Up}\label{subsec:experimental_setup}
We investigate a car-like robot where the discrete-time dynamics are given by applying a zero-order hold on controls and disturbance for the following kinematic bicycle model with time step $\Delta t = 0.5$ seconds,
\begin{equation*}
    \begin{split}
        &\dot{x} = V \cos(\psi + \beta),\quad \dot{y} = V \sin(\psi + \beta),\quad
        \dot{\psi} = \frac{V}{l_\mathrm{R}} \sin(\beta), \\
        &\dot{V} = a + d_a, \qquad 
        \tan(\beta) = \frac{l_\mathrm{R}\tan(\delta + d_\delta )}{l_\mathrm{R} + l_\mathrm{F}}.
    \end{split}
    \label{eq:kinematic bicycle}
\end{equation*}
The speed of the vehicle is bounded, $V \in [0, 5]$ (m/s), the controls are bounded with $a \in [-3, 3]$ (ms$^{-2}$) and $\delta \in [-0.344, 0.344]$ (radians), and the distance from the center of mass to the front and rear axles are $l_\mathrm{F}=0.5$m, and $l_\mathrm{R}=0.7$m respectively. There is a disturbance $d=[d_\delta, d_a]$ applied onto the control inputs with $d \sim \mathcal{N}(\mu=[0,0], \Sigma=\mathrm{diag}([0.05, 0.02]))$. We set $T=55$.

The environment is characterized by an initial state, coverage, obstacle, and goal region as shown in Figure~\ref{fig:human trajectory} (left). The image $e$ represents a top-down view of the environment.\footnote{The shape and position of the regions could be parameterized with a vector of numbers instead, but we use images to illustrate the generality of our approach.}
The position of the coverage (white circle) and obstacle (red circle) region can vary---the $x$-position of the coverage region, $x_\mathrm{cov}$, varies with a fixed $y$-position, while the $x$-position of the obstacle is always half way between the coverage and goal region.
We constrain the environment in this way to ensure the problem remains feasible for a fixed time horizon and avoids instances where it is trivial to avoid the obstacle region.
The regions are assumed to be circles to ease STL predicate computations, but in general can be more complex as long as we can backpropagate through the STL predicates.
We consider the following STL specification,
\begin{equation}
    \begin{split}
    \phi &= ((\lozenge \square_{[0,8]}\, \phi_\mathrm{cov}) \;\mathcal{U}\; (\lozenge \,\phi_\mathrm{goal})) \, \wedge \, \square\, \neg \phi_\mathrm{obs}\\
        \phi_\mathrm{cov} &=\mu_\mathrm{cov}(x_t) < 0 \wedge V_t < 2.0\\
        \phi_\mathrm{goal} &= \square \, (\mu_\mathrm{goal}(x_t) < 0 \wedge V_t < 0.5)\\
        \phi_\mathrm{obs} &= \mu_\mathrm{obs}(x_t) < 0.
    \end{split}
    \label{eq:phi details}
\end{equation}
In words: \emph{$\phi$ requires the robot to first slow down to less than 2m/s inside the coverage region for $8\Delta t$ seconds before moving into the goal region and staying inside with velocity less than 0.5m/s. Simultaneously, the robot should always avoid entering the obstacle region.}
We highlight that \eqref{eq:phi details} is more complex than the reach-avoid specifications studied in related works \cite{YaghoubiFainekos2019,LiuMehdipourEtAl2021} because (i) \eqref{eq:phi details} consists of a bounded time interval indicating the minimum duration to stay inside a coverage region, (ii) there are restrictions on the velocity of the robot, and (iii) there are three nested temporal operators whereas others have at most two.

We provide 32 expert demonstrations satisfying $\phi$ which were collected in simulation with a human using an XBox controller to control the robot (see Figure~\ref{fig:hero}).
The simulation environment was implemented using the Robot Operating System (ROS) and visualized in RViz.
We used PyTorch \cite{PaszkeGrossEtAl2017} to implement our neural network controller, and \texttt{stlcg} \cite{LeungArechigaEtAl2020} for the STL robustness calculations. 
% The code accompanying this work can be found at \url{https://github.com/StanfordASL/stlcg_ctrl_syn}.
% More details about the neural network parameters and hyperparameters used for the training process can be found in Appendix \ref{app:neural network parameters} and Appendix \ref{app:hyperparameters} respectively.

\subsection{Analysis and Discussion}
We first discuss the offline training procedure, and then the performance of our proposed online adaptive method including comparisons to a baseline approach.

\subsubsection{Offline training}
Figure~\ref{fig:human trajectory} illustrates the closed-loop trajectories (without the online adaptation step) produced by $\pi_\theta$ trained with different values of $\gamma$, the weighting on the imitation loss. 
Interestingly, when $\gamma=0$, the case where we optimize over STL robustness only, the controller performs \emph{worse}---the controller converges to a local optimum which produces trajectories that pass only to the right of the obstacle and is therefore unable to reach the coverage region whenever the coverage region is to the left (see second plot in Figure~\ref{fig:human trajectory}).
When $\gamma >0$, the model is able to mimic the expert trajectory and pass to the left or right of the obstacle depending on the environment configuration.
This behavior indicates that even though the primary goal is to satisfy $\phi$, optimizing only for STL robustness is not the most effective as it can very easily converge to a clearly sub-optimal and non-intuitive solution.
Instead, simply using a few expert demonstrations can guide the policy exploration to a better local optimum---one achieves better STL satisfaction and also mirrors naturalistic and intuitive behaviors.

\begin{figure}[t]
    \centering
    \includegraphics[width=0.48\textwidth]{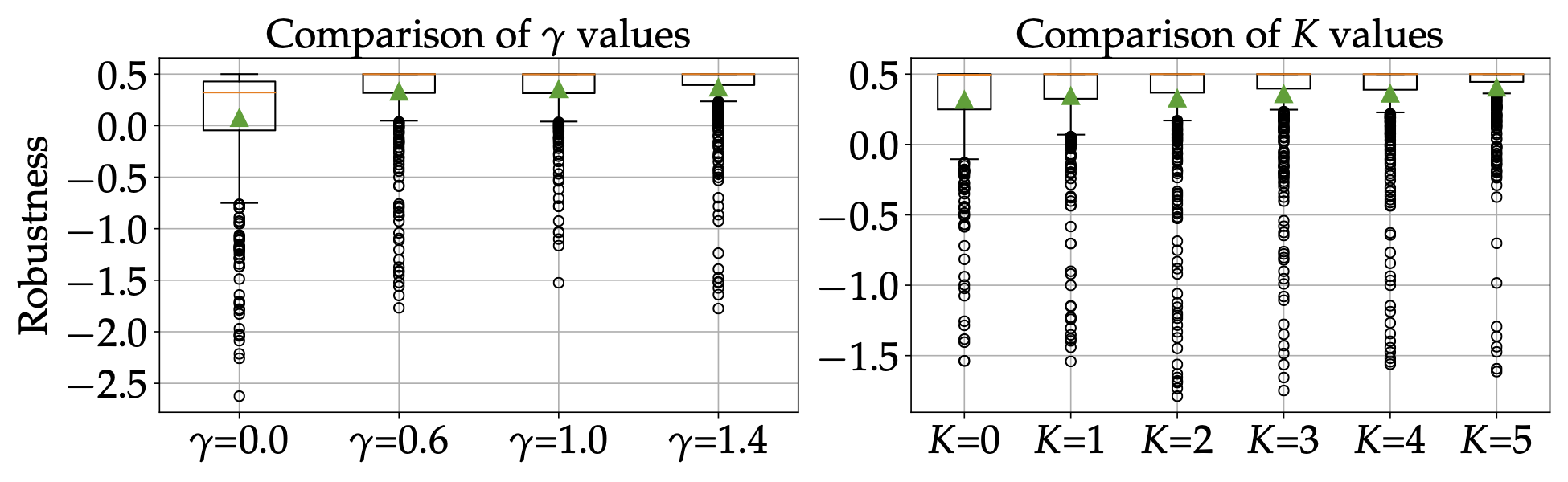}
    \caption{Distribution of robustness values (512 samples) using policies trained with different $\gamma$ values (left) and adversarial training iterations $K$ (right). No online adaptation is used in computing the robustness value. Orange bar indicates the median, and the green triangle denotes the mean.}
    \label{fig:robustness gamma K distribution}
\end{figure}

For a quantitative comparison, the distribution of robustness values when using models trained with different $\gamma$ values is presented in Figure~\ref{fig:robustness gamma K distribution} (left). The mean and median is the highest when $\gamma=1.4$.
Figure~\ref{fig:robustness gamma K distribution} (right) illustrates the STL robustness distribution corresponding to controllers trained with different numbers of adversarial training iterations ($K$) and with $\gamma = 1.4$.
We see that using more adversarial steps help shift the distribution towards higher robustness.
Moving forward, we will use a model trained with $\gamma=1.4$ and $K=5$ in the following results.

\subsubsection{Online performance}
We compare both qualitatively and quantitatively the performance of our proposed trajectory-feedback STL controller (with and without online adaptation) against a baseline shooting method adapted from \cite{PantAbbasEtAl2017}. Given the nonlinear dynamics, problem size, and specification complexity, MILP approaches \cite{RamanDonzeEtAl2014} would not be a suitable comparison. Here, we describe the different controllers that we investigate.

\noindent {\bf Baseline}: Consider an optimal control problem,
\begin{equation}
    \begin{split}
        \ut{t:T-1}^* &= \max_{\ut{t:T-1}} \, \rho_\phi(\xi_{\tau_t, T}^{\ut{t:T}, 0})\\
        \text{s.t.} \quad & \underline{u} \leq \ut{t^\prime} \leq \overline{u}, \quad \tau_t = \xt{0:t}\\ 
        & \xt{t^\prime+1} = f(\xt{t^\prime}, \ut{t^\prime}, 0), \quad  \forall t^\prime=t,...,T-1,
    \end{split}
    \label{eq:baseline opt}
\end{equation}
where the system has past trajectory $\xt{0:t}$. Note the zero disturbance in the dynamics.
We used a projected limited-memory BFGS (L-BFGS) \cite{ByrdNocedalEtAl1994} gradient descent optimizer to solve \eqref{eq:baseline opt}. We used PyTorch's built-in L-BFGS optimizer with step size 0.05 and clipped $\ut{t}$ to make sure the control constraints were satisfied.
Similar to Algorithm~\ref{alg:online adaptation}, the optimization is only performed if the planned trajectory results in negative robustness. This shooting method is designed to mimic the SQP approach proposed in \cite{PantAbbasEtAl2017}. Since the Hessian computation took $\sim10$ seconds in PyTorch, we opted for a quasi-newton method instead to reduce computation times. Due to the highly nonlinear nature of \eqref{eq:baseline opt}, we use the control sequence generated by propagating $\pi_\theta$ to warm-start \eqref{eq:baseline opt} at the first time step, and the solution from the previous time step thereafter. We set $N_\mathrm{gd}=3$, the maximum number of gradient steps at each time step.

\noindent {\bf Open-loop}: Given the initial state and environment, we compute the control sequence resulting from propagating the state over the time horizon using $\pi_\theta$. Then we execute the control sequence in an open-loop fashion.

\noindent {\bf Trajectory-feedback (TF)}: At each time step, the past trajectory is passed into $\pi_\theta$ to compute the next control input. No online gradient steps will be used in this approach.
    
\noindent {\bf Trajectory-feedback with online adaptation (TF$^*$--$N_\mathrm{gd}$)}: The TF approach with online adaptation. This represents the core approach proposed in this paper. We consider two cases, $N_\mathrm{gd} = 1$ (TF$^*$--1) and $N_\mathrm{gd}=3$ (TF$^*$--3). We use the default Adam optimizer \cite{KingmaBa2015} in PyTorch.

\begin{figure}[t]
    \centering
    \includegraphics[width=0.4\textwidth]{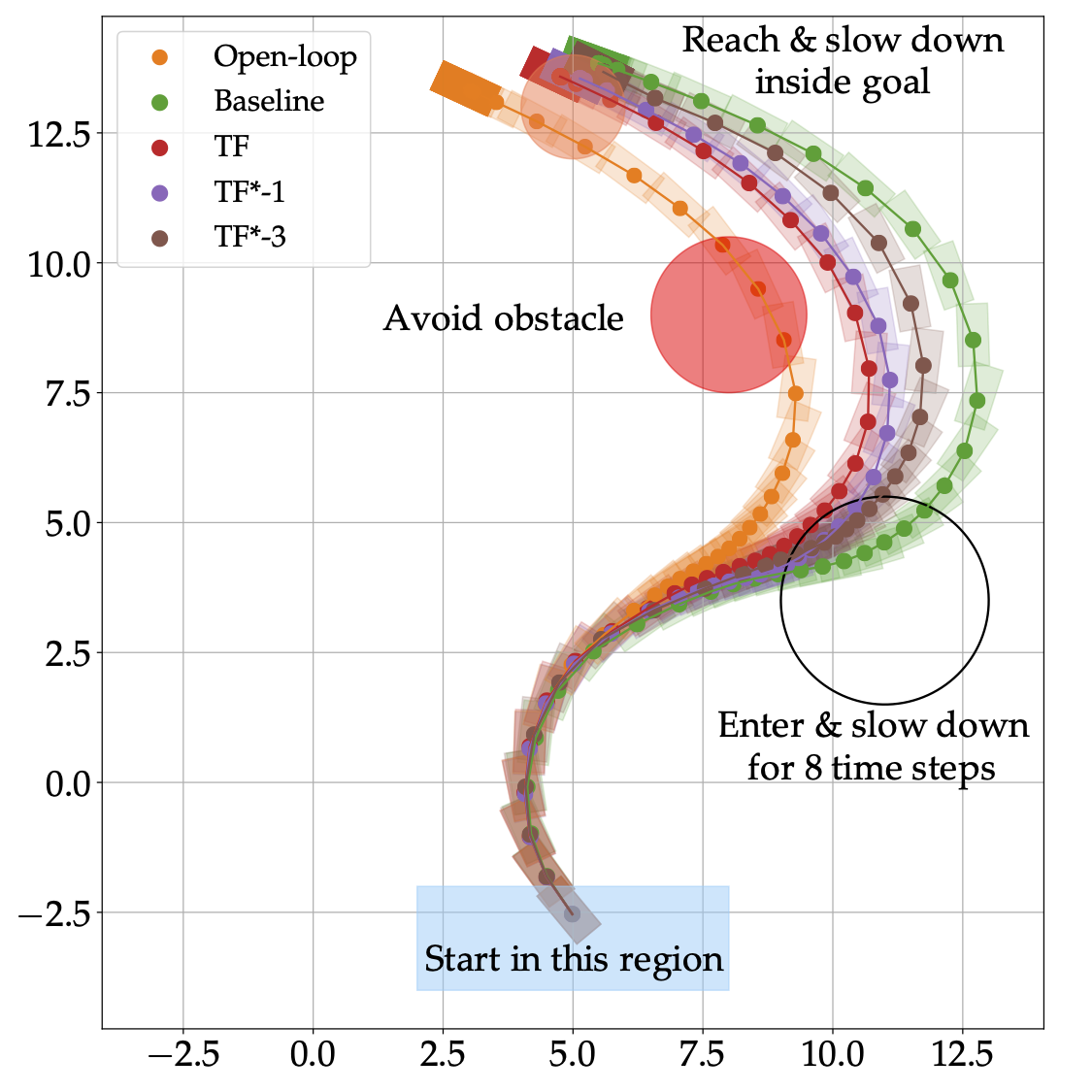}
    \caption{Trajectories generated from different online control strategies. Our proposed method (TF$^*$) satisfies the STL specification and achieves the highest robustness. The corresponding robustness values are (higher is better): Open-loop: -1.08, Baseline: 0.27, TF: -0.29, TF$^*$--1: 0.13, {\bf TF$^*$--3: 0.42}.}
    \label{fig:trajectories_deploy}
\end{figure}

We simulated (with noise) each control strategy, and Figure~\ref{fig:trajectories_deploy} showcases the trajectories deployed from each of the aforementioned methods for a particular environment. 
To better highlight the features of each method, we chose a challenging environment with $x_\mathrm{cov}=11$ which is outside of the distribution used to generate training data, $x_\mathrm{cov} \sim \mathrm{Uniform}[1, 10]$.
The open-loop and TF approaches resulted in negative robustness, while the Baseline and TF$^*$ approaches, both of which are able to adjust to new environments, resulted in positive robustness values with TF$^*$--3 achieving the highest value. These behaviors highlight the significance of the online gradient steps in producing satisfying trajectories.

\begin{figure}[t]
    \centering
    \includegraphics[width=0.48\textwidth]{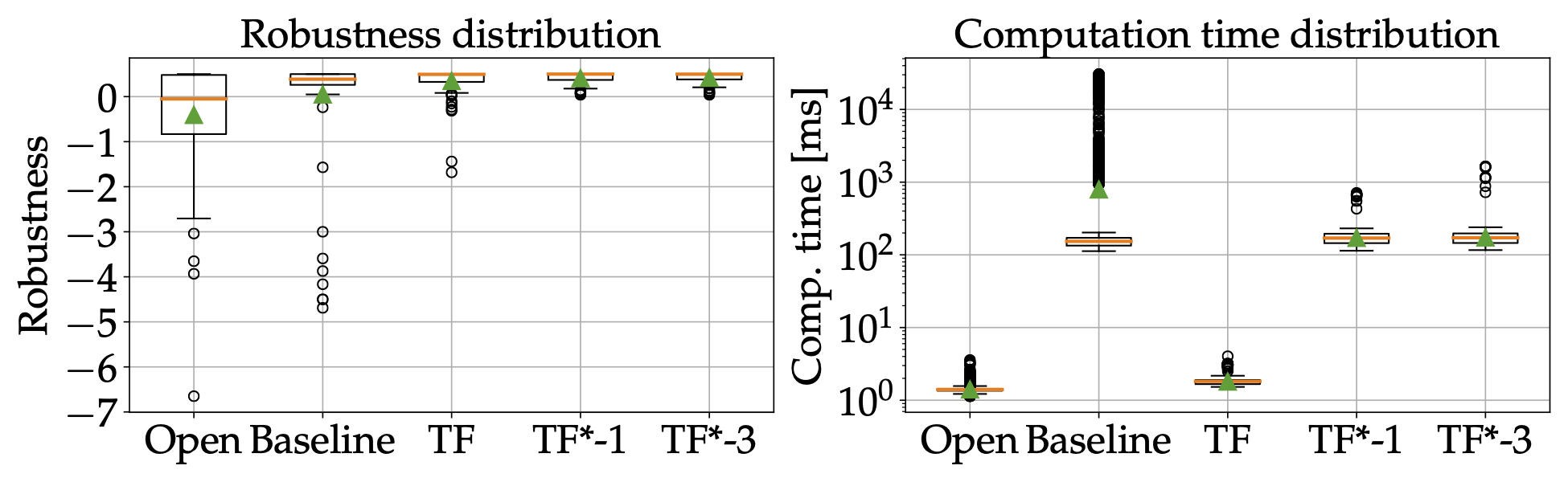}
    \caption{Comparison of robustness (left) and computation time to compute a control input at each time step (right). 100 random initial states and environments were used. Orange bar indicates the median, and the green triangle denotes the mean. The simulations were ran using a 3.0GHz octocore AMD Ryzen 1700 CPU and a Titan X (Pascal) GPU.}
    \label{fig:robustness computation random time}
\end{figure}

We also ran 100 trials with random initial states and environments (consistent with the training distribution) and compared the STL robustness values and computations (see Figure~\ref{fig:robustness computation random time}).
We highlight four takeaways from these results: (i) Rolling out the trajectory over the entire time horizon and evaluating the robustness value takes roughly 100ms. 
The computation time may be reduced with more tailored software.
(ii) Both TF$^*$--1 and TF$^*$--3 produces 100\% success rate in producing satisfying trajectories with TF$^*$--3 having a slightly higher mean robustness. However, in terms of computation times, there were a few instances (less than 0.5\%) where the computation time was greater than 300ms, corresponding to times where gradient steps were needed. 
With more tailored software, there is high potential for the computation time to reach real-time applicability.
(iii) Baseline has a 93\% success rate, but about 10\% of the time steps result in a computation time of roughly 1000ms or more, making it difficult to reach real-time applicability even with tailored software.
(iv) As expected, the Open-loop and TF approaches have very low computation times. The success rates are 48\% and 91\% respectively. As such, the TF and Baseline methods perform similarly in terms of robustness performance, but TF is more desirable due to its lower computation time.

\subsection{Autonomous driving example}
\begin{figure}
    \centering
    \includegraphics[width=0.45\textwidth]{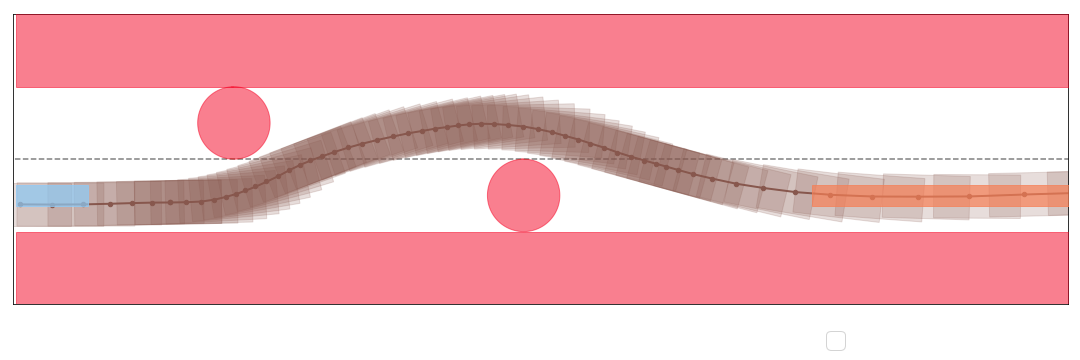}
    \caption{Trajectory of a car-like robot satisfying STL specification $\psi$. Obstacles, initial states set, and goal set are denoted by the pink, blue and red regions respectively.}
    \label{fig:driving example}
\end{figure}

We synthesized another controller for an autonomous driving setting using a new STL specification reminiscent of a car approaching a road construction site,
\begin{equation*}
    \begin{split}
        \psi &= \psi_\mathrm{obs} \: \wedge \: \psi_\mathrm{slow} \: \wedge \: \psi_\mathrm{goal},\\
        \psi_\mathrm{obs} &= \square\, (\text{avoid road boundary} \, \wedge \, \text{avoid obstacles}),\\
        \psi_\mathrm{slow} &= \square \, (\psi_\mathrm{near} \rightarrow V < 0.55), \, \psi_\mathrm{near} = \text{dist. to obs.} < 1.2,\\
        \psi_\mathrm{goal} & = \lozenge \, \square _{[0,2]} (\text{in goal} \: \wedge \: V > 1.0).
    \end{split}
\end{equation*} 
The specification $\psi$ requires a car to stay on the road and avoid obstacles, slow down when near obstacles, and speed up for at least 2 time steps once it is in the goal region. The positions of the obstacles can vary. 
We synthesized a new controller with $\psi$ using eight human demonstrations and
Figure~\ref{fig:driving example} illustrates a trajectory computed with $N_\mathrm{gd}=3$.

\section{Conclusions and Future Work}\label{sec:conclusions_future_work}
We have presented a semi-supervised approach for synthesizing a trajectory-feedback controller designed to satisfy a desired STL specification in varied environments.
By utilizing few expert demonstrations for training regularization and an online adaptation phase, the controller consistently satisfied the desired STL specification while maintaining naturalistic human behaviors. 
Although verifying spatio-temporal properties of closed-loop neural network controllers remains an open problem, we showed that an online adaptation phase is significant in bolstering the spatio-temporal performance of a neural network controller.
We showed through an illustrative case study that our proposed controller achieves better performance and lower computation times compared to a shooting method adapted from a state-of-the-art approach.
Future work includes (i) learning a value function to avoid long roll outs and therefore reduce the computation time, (ii) using more complex dynamics (e.g., expressed as neural networks) and environment representation such as a map or camera images attached to the robot, and (iii) extending the controller to account for multi-agent settings where STL specifications become even more complex.

\bibliographystyle{IEEEtran}
% \bibliography{IEEEabrv,ASL_papers,main}
\bibliography{../../../bib/ASL_papers,../../../bib/main}

\end{document}